\documentclass[letterpaper, 10 pt, conference]{ieeeconf}  % Comment this line out if you need a4paper
\usepackage{multirow}
\IEEEoverridecommandlockouts          
\usepackage{graphics} % for pdf, bitmapped graphics files
\usepackage{epsfig} % for postscript graphics files
\usepackage{mathptmx} % assumes new font selection scheme installed
\usepackage{times} % assumes new font selection scheme installed
\usepackage{amssymb}  % assumes amsmath package installed
\usepackage{graphicx}
\usepackage{algorithm}
\usepackage{algorithmic}
\usepackage{caption}
\usepackage{mathtools}
\usepackage{mathrsfs, amsmath} 
\usepackage{color}
\usepackage{xcolor}

\usepackage{array}
\newcolumntype{P}[1]{>{\centering\arraybackslash}p{#1}}
\newcolumntype{M}[1]{>{\centering\arraybackslash}m{#1}}

\title{\LARGE \bf
CAROM - Vehicle Localization and Traffic Scene Reconstruction from Monocular Cameras on Road Infrastructures
}

\author{Duo Lu$^{1}$, Varun C Jammula$^{1}$, Steven Como$^{1}$, Jeffrey Wishart$^{2}$,  Yan Chen$^{1}$, Yezhou Yang$^{1}$%
\thanks{$^{1}$D. Lu, V. Jammula, S. Como, Y. Chen and Y. Yang are with Arizona State University, Tempe, AZ, USA. {\tt\small \{duolu, vjammula, scomo, yanchen, yz.yang\}@asu.edu}}%
\thanks{$^{2}$ J. Wishart is with Exponent, Tempe, AZ, USA. {\tt\small jwishart@exponent.com}}%
\thanks{This research is sponsored by the Institute of Automated Mobility, Arizona, USA. We thank Maricopa County DOT, Niraj Vasant Altekar, Larry Head, Alex Cardona, Don Bruyere, Maria Elli, Jack Weast, Greg Leeming, and Marisa Paula Walker for their help.}
}

\begin{document}

\maketitle
% \thispagestyle{empty}
% \pagestyle{empty}

%%%%%%%%%%%%%%%%%%%%%%%%%%%%%%%%%%%%%%%%%%%%%%%%%%%%%%%%%%%%%%%%%%%%%%%%%%%%%%%%

\begin{abstract}

Traffic monitoring cameras are powerful tools for traffic management and essential components of intelligent road infrastructure systems. In this paper, we present a vehicle localization and traffic scene reconstruction framework using these cameras, dubbed as CAROM, \textit{i.e.}, ``\underline{CAR}s \underline{O}n the \underline{M}ap''. CAROM processes traffic monitoring videos and converts them to anonymous data structures of vehicle type, 3D shape, position, and velocity for traffic scene reconstruction and replay. Through collaborating with a local department of transportation in the United States, we constructed a benchmarking dataset containing GPS data, roadside camera videos, and drone videos to validate the vehicle tracking results. On average, the localization error is approximately 0.8 m and 1.7 m within the range of 50 m and 120 m from the cameras, respectively.

%, and thus set up a standard evaluation protocol for future research. On average, the localization error is within 2 m and the speed measurement error is less than 1.5 m/s. %To demonstrate the usability of the system, we conduct case studies on a critical civil task: automatic vehicle safety metric estimation. 

\end{abstract}

\section{Introduction}

Traffic monitoring cameras and smart roadside units with vision-based sensors are becoming increasingly popular for traffic management purposes. Local Departments of Transportation (DOTs) use the videos to investigate driving safety, study traffic congestion, and sometimes issue tickets for rule violations. This type of equipment is also an essential component of the intelligent road infrastructure system for the ``automated vehicles'' in the future. However, there are three unsolved problems related to these cameras. First, transmitting and archiving the videos cost a significant amount of network bandwidth and storage space. Second, these videos are unfriendly to index, search, and automated analysis since they contain mainly unstructured information. Especially, it is difficult to obtain 3D states of the vehicles from 2D images. For example, local DOTs often require traffic management officers to monitor and interpret the videos to evaluate driving safety based on safety-critical events such as accidents. These events do not happen very frequently. The analysis can also involve subjective bias, e.g., errors in determining the velocity of vehicles or distance between vehicles due to the restriction of camera perspective angles. Third, privacy concerns haunt the civil usage of these videos, which restricts non-authority organizations to access them. For example, it is usually not acceptable for a local DOT to send the raw videos to third-party companies or scholars for data analysis unless they are contracted by the DOT. Also, insurance companies cannot access them for improving the process of traffic incident claims. 

\begin{figure}[]
    % \centering
    \begin{center}
        \includegraphics[width=3.4in]{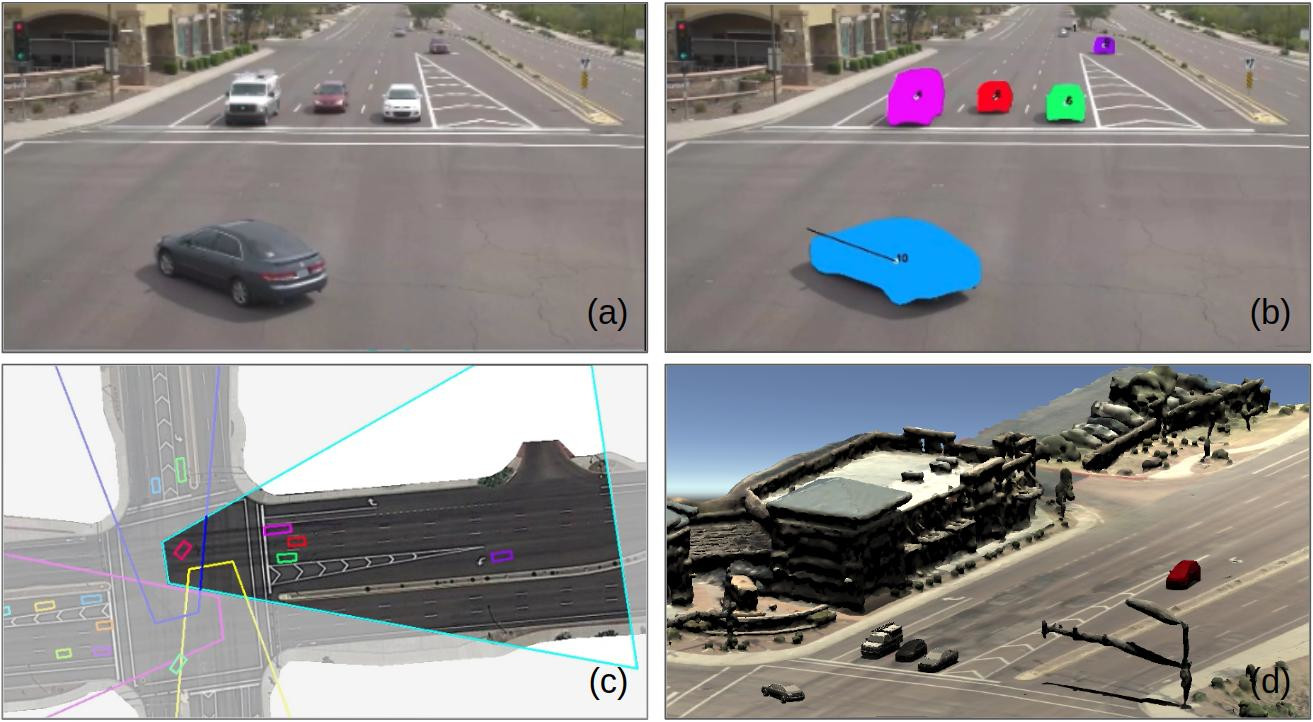}
    \caption{An overview of CAROM: (a) original traffic monitoring video, (b) detected vehicles, (c) replay on a 2D map, (d) replay on a 3D map.}
    \label{fig:overview}
    \end{center}
\vspace{-0.25in}
\end{figure}

Automated vehicles (AVs) can also benefit from these cameras. For example, it is a tough question for both the DOTs and the manufacturers to answer how safe are the AVs currently testing on the road. The traffic monitoring cameras are commonly mounted on road infrastructures with the advantage of covering a large area. Hence, they can be used to objectively assess the operational safety by calculating a set of safety metrics \cite{wishart2020driving} directly from vehicle movements captured on the videos. Meanwhile, the information of the surrounding traffic scene obtained by these cameras can complement the perception of AVs because the in-vehicle sensors can only reach places in the line-of-sight.

To address these issues, we propose CAROM, a framework that can extract 3D information from the videos, generate a series of structured data records of vehicle states, and reconstruct traffic scenes on a 2D map or a 3D map, as shown in Fig \ref{fig:overview}. This work is part of research being conducted by the Institute of Automated Mobility (IAM) \cite{IAM} to develop an operational safety assessment methodology and an intelligent automated infrastructure for vehicles. CAROM facilitates a series of applications including road safety evaluation, roadside information services for AVs, traffic data archiving, sharing, and further automated analysis. The generated data records can be saved in a database or sent over the network with significantly less storage and bandwidth cost than the raw videos. Moreover, a reconstructed traffic scene can be replayed to offer an objective vision of the traffic situations in a bird's-eye view to an interested organization besides the video owner. Last but not the least, the generated data could be easily anonymized by removing any Personally Identifiable Information (PII). In summary, our contributions are as follows:

\begin{enumerate}

\item We constructed a vehicle tracking, localization, and velocity measurement pipeline using videos taken by monocular road traffic monitoring cameras.

\item We built a reconstruction system for vehicle shapes and traffic scenes using the tracking results. Additionally, we created two visualizers to replay the reconstructed traffic scene on both 2D and 3D maps.

\item We evaluated the vehicle localization and velocity measurement performance using both differential GPS and drone videos, which shows promising results.

\end{enumerate}

%Although the current implementation records and processes videos in batches,  can be deployed on public roads together with the cameras in the future to help regulators in local DOTs and make future self-driving vehicles safer on the road.

%The remainder of this paper is organized as follows. Section 2 describes the related works, and section 3 explains the details of the CAROM framework. Section 4 provides empirical performance evaluation results with discussions of limitations. Finally, section 5 draws the conclusions.

\section{Related Work}

With the advancement of effective neural network object detectors \cite{ren2015faster}\cite{he2017mask}, tracking algorithms \cite{wojke2017simple}, and large scale datasets \cite{geiger2013vision}\cite{naphade20204th}, current research work has obtained great achievements in video based road traffic analysis in the past decade \cite{gupte2002detection}\cite{sivaraman2013looking}\cite{liu2013survey}\cite{kumaran2019anomaly}\cite{tian2011video}\cite{datondji2016survey}. Commercial video analysis software platforms as well as roadside smart cameras are also emerging \cite{Transoft}\cite{NoTraffic}. However, localization of vehicles, speed measurement, and reconstruction of traffic scenes in 3D space are still challenging due to two core problems. First, accurate calibration of the cameras is necessary to convert 2D pixels to 3D locations. This can be done manually using labeled point correspondences or automatically using vanishing points calculated from geometric primitives \cite{dubska2014automatic}\cite{corral2014automatic}\cite{lee2011robust} and objects with known shapes \cite{sochor2017traffic}. Typically, the automated calibration algorithm also sets up a 3D world reference frame (not related to any predefined map). Second, in addition to accurate vehicle detection and tracking on the 2D images, robust estimation of vehicle 3D pose and vehicle dimension is required. The 3D representation of a vehicle can be a point with an orientation vector \cite{juranek2015real}, a 3D bounding box \cite{dubska2014automatic}\cite{sochor2016boxcars}\cite{mousavian20173d}, a few key points \cite{zhang2020vehicle}, a wireframe model \cite{ding2018vehicle}\cite{ansari2018earth}, or a parametric 3D shape model \cite{leotta2010vehicle}. The location and speed of a vehicle are typically determined from three pieces of information: (1) the 2D locations on the images, (2) the 3D poses of the vehicle, and (3) the transformation between image coordinates and the ground coordinates obtained from the camera calibration results. Usually, the vehicle states are also estimated jointly through a filtering process by considering the vehicle kinematics or dynamics \cite{chen2011kalman}\cite{li2018generic}. The vehicle shape can be reconstructed using stereo cameras \cite{engelmann2017samp} or monocular cameras \cite{ansari2018earth}\cite{chhaya2016monocular}\cite{prisacariu2012simultaneous} through a sequence of algorithms for depth estimation, model fitting, and shape optimization. Our paper is based on the existing works for several individual computer vision tasks and we integrated them to a unified framework that extracts the location, speed, and vehicle shape in the 3D space. Further, our tracking results allow the vehicle movements to be replayed on a 2D map or a 3D map so as to support traffic analysis tasks.

The use of simulators with a single vehicle or a collection of vehicles have been studied intensively to visualize vehicle motion, study vehicle dynamics, understand traffic patterns, and train driving behaviors of AVs \cite{dosovitskiy2017carla}\cite{lopez2018microscopic}. Unlike these works, we desire to ``re-simulate'' the traffic scenes using the reconstruction results from the videos.

\section {The CAROM Framework Architecture}

\begin{figure}[ht]
    % \centering
    \begin{center}
        \includegraphics[width=3.2in]{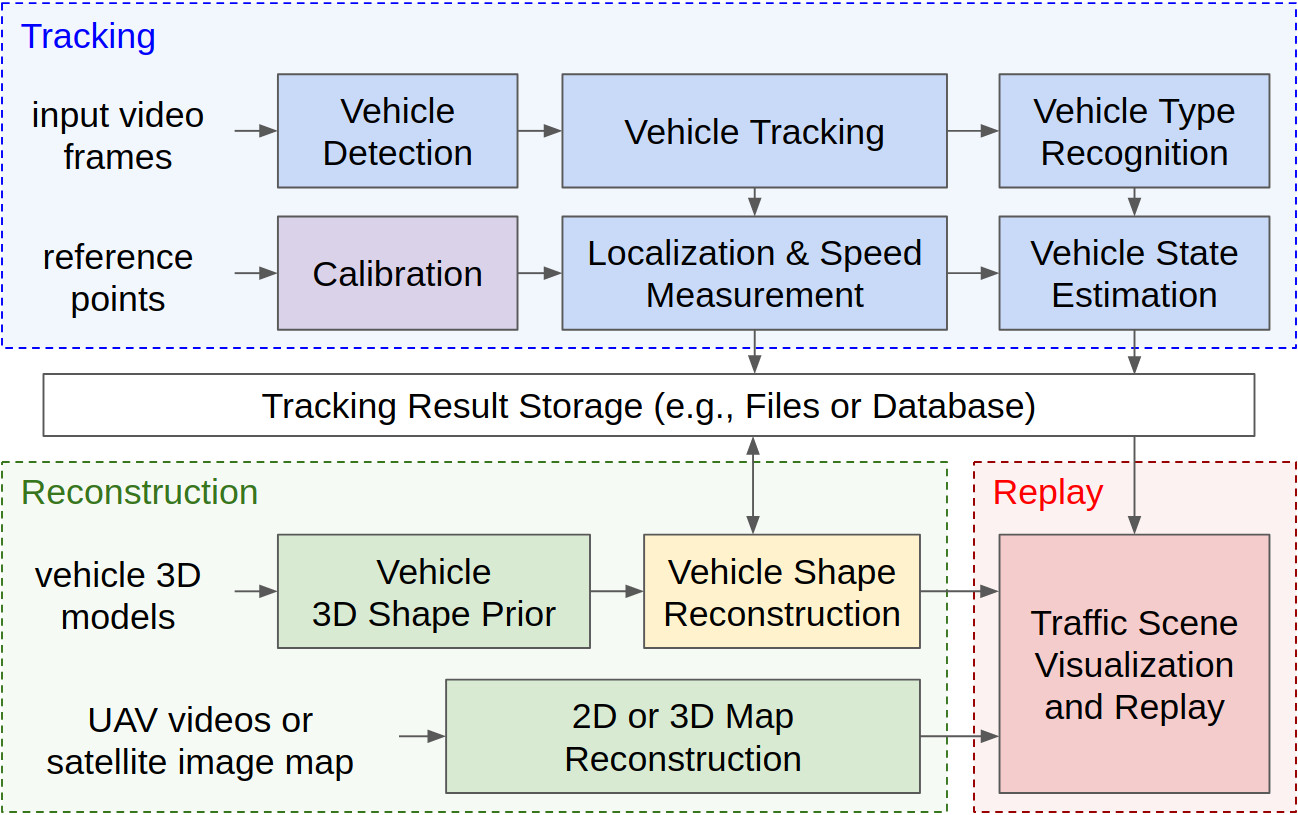}
    \caption{The CAROM Framework Architecture}
    \label{fig:arch}
    \end{center}
\vspace{-0.2in}
\end{figure}

The CAROM framework consists of three subsystems, as illustrated in Fig \ref{fig:arch}. The first one is the tracking system, which runs a pipeline to generate data structures of vehicle states from videos. This pipeline contains an offline calibration stage (detailed in Section III.A) and a few online video processing stages, including vehicle detection, tracking, localization, type recognition, and 3D state estimation (detailed in Section III.B). The generated data structures can be stored in files or a database for future usage, such as road safety assessment. The second one is the reconstruction system for vehicle shapes (detailed in Section III.C) and the map (detailed in Section III.D). 
%The vehicle reconstruction module takes a sequence of instance segmentation masks, the set of corresponding vehicle states from the tracking results, and a vehicle 3D shape prior to compute a vector representation of the shape of the reconstructed vehicle (detailed in section III.C). The 3D map reconstruction module uses a SLAM software to create a 3D mesh map of a road segment from data collected by a drone (detailed in section III.D). If the road segment is mostly flat and a 3D map is not necessary, a 2D satellite image map can be used instead, which is generally easier to reconstruct. 
The third subsystem is the replay engine that animates the traffic scene on the reconstructed map using the tracking results (detailed in Section III.E).

\subsection{Camera and Map Calibration}

\begin{figure}[]
\vspace{0.07in}
    % \centering
    \begin{center}
        \includegraphics[width=3.0in]{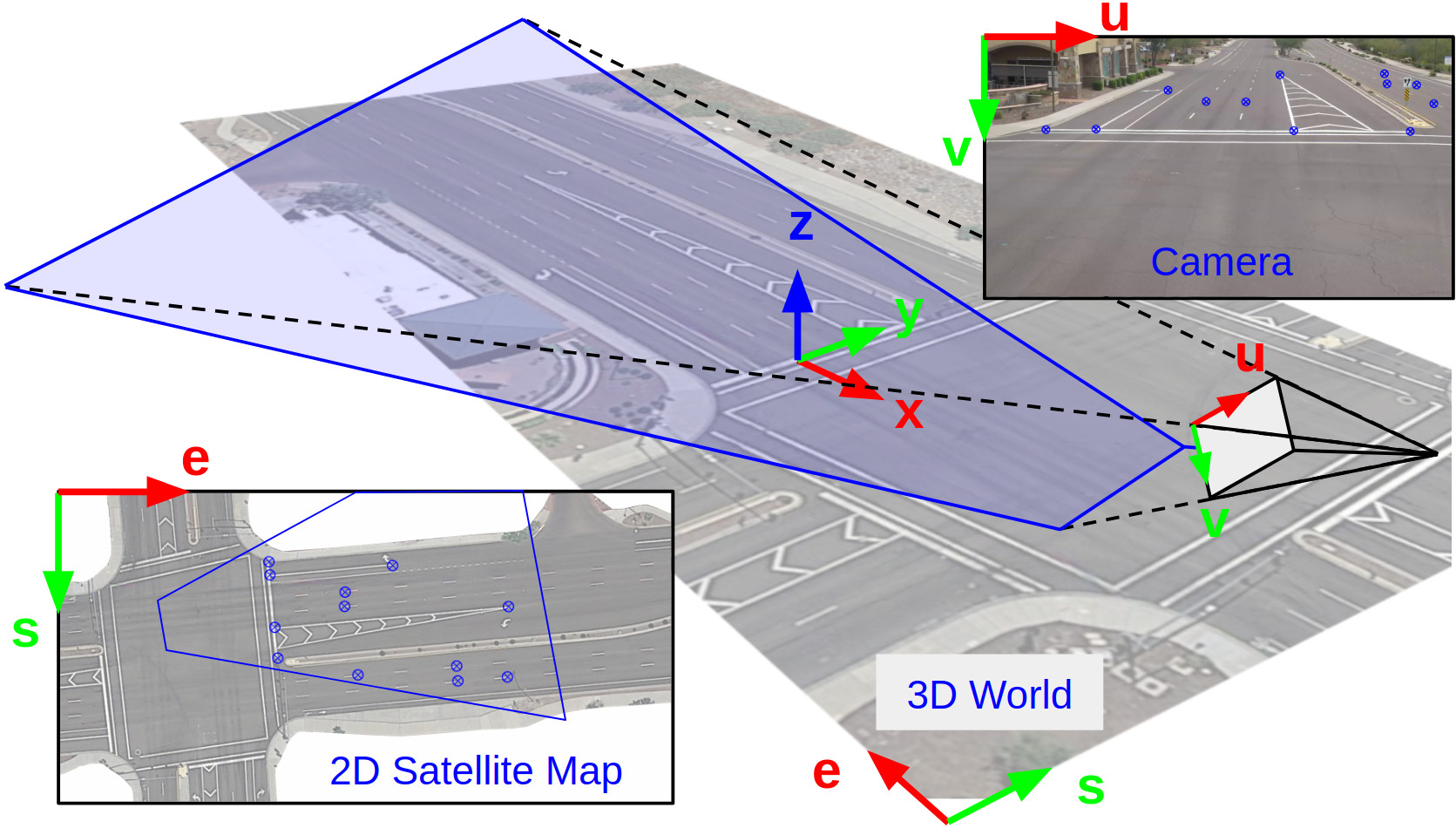}
    \caption{An illustration of reference frames and point correspondences.}
    \label{fig:calibration}
    \end{center}
\vspace{-0.2in}
\end{figure}

\begin{figure}[]
    % \centering
    \begin{center}
        \includegraphics[width=3.0in]{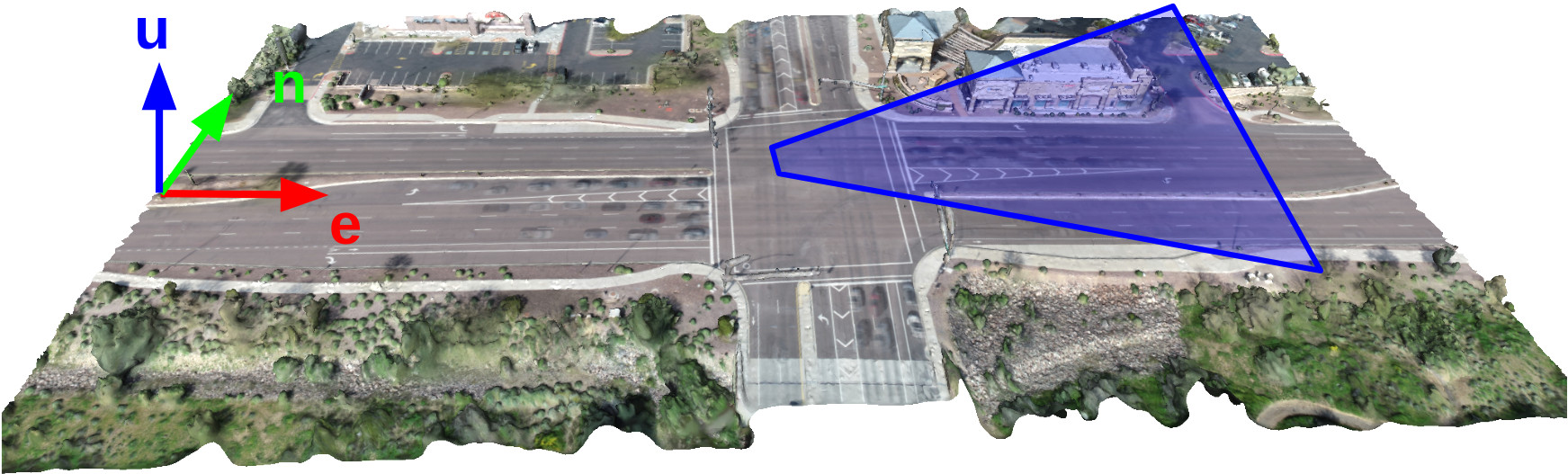}
    \caption{An example of the 3D map and camera coverage.}
    \label{fig:map_3d}
    \end{center}
\vspace{-0.3in}
\end{figure}

CAROM uses a pinhole camera model and assumes the camera distortion is negligible, as illustrated in Fig. \ref{fig:calibration}. The ground is modeled either as a flat surface corresponding to a 2D satellite image map or a 3D surface with a high-resolution 3D mesh map, as shown in Fig. \ref{fig:map_3d}. There are three reference frames: (1) the camera frame in image pixel coordinates, (2) the world frame in metric coordinates, and (3) the map frame in map coordinates. For a 2D map, the origin is the top-left corner, the axes follow east-south directions, and the unit is a map pixel (as in Fig. \ref{fig:calibration}). For a 3D map, the origin can be any point on the ground surface, the axes follow east-north-up directions, and the coordinates use the metric unit (as in Fig. \ref{fig:map_3d}). The calibration procedure constructs two sets of parameters: (1) a camera projection matrix from the world frame to the camera frame, (2) a transformation between the map frame and the world frame. Since the traffic monitoring cameras do not move, we only need to run the calibration procedure once for each camera in the following steps. First, we label a set of at least six point correspondences on the map and the image, typically using the lane markers and features on the ground. Second, we create the world frame and compute the transformation between the world frame to the map frame. Usually, the XOY plane of the world frame is the ground plane in the 2D map and the x-axis follows the traffic moving direction. Third, we transform the labeled points on the map to the world frame and compute the camera projection matrix from the point correspondences \cite{hartley2003multiple}. Optionally, the calibration of the camera can be automated \cite{dubska2014automatic}, but the transformation between the world frame and the map still needs to be determined using labeled point correspondences. The transformation from any image coordinates to the world frame on the ground is crucial for vehicle localization, and we denoted it as $T$. If a 2D map is used, $T$ is the planar homography between the camera frame and the ground plane, which is derived from the camera projection matrix. If the 3D map is used, we back-project each pixel on the image to the 3D ground surface to obtain its corresponding point in the world frame. Then we construct $T$ as a look-up-table. Additionally, we also compute the horizon line on the image from the camera projection matrix.

\subsection{Online Vehicle Tracking Pipeline}

The tracking pipeline consists of a set of online algorithms that independently processes every image in a video. It uses information from the previous images for vehicle speed measurement. With enough processing resources, it may be able to run in real-time. It has the following stages.

%\textbf{(1) Image Stabilization}: A traffic monitoring camera is usually mounted on a traffic light pole or other pole-shaped the road infrastructure which can shake with the wind. To stabilize the video, we first compute the optical flow \cite{lucas1981iterative} between the current image and the reference image used in the camera calibration. Then, we estimate a homography between these frames and warp the current image to remove the shake. This step is optional if the camera does not shake.

\textbf{(1) Vehicle Detection}: For each video image, the system runs an object detection and instance segmentation network. In our implementation, We fine-tuned a Mask RCNN \cite{he2017mask} on a custom dataset created from traffic monitoring videos for this step. The quality of the masks is crucial since the later localization stage relies on the contour of the mask.

\textbf{(2) Vehicle Tracking}: For each detected object instance, its 2D bounding box on the current image is enlarged four times as a region-of-interest (ROI). The sparse optical flow vectors \cite{lucas1981iterative} from the previous image to the current image are calculated within this ROI and on the masks. The detected instances on the two images are associated in linked lists using these vectors and the mask overlapping percentages.

\textbf{(3) Vehicle Type Recognition}: For each detected object instance, the system crops a square patch from the image just large enough to contain its 2D bounding box, resizes the cropped patch and runs a classifier to predict its type. The following types are used: $\{$pedestrian, two-wheelers, bus, mini-truck, semi-truck, pickup-truck, convertible, coupe, sedan, all-terrain vehicle, minivan, van, SUV, trailer$\}$. In our implementation, we trained a ResNet-18 \cite{he2016deep} on a custom dataset for this step. The decoupling of detector and vehicle type classifier is intentional, which makes both neural networks easier to train. Since this recognizer can learn a different set of features dedicated to its task regardless of the detector, it may also perform better. Moreover, we plan to build a fine-grained vehicle make and model classifier to replace this vehicle type recognizer in the future.

\begin{figure}[]
\vspace{0.1in}
    % \centering
    \begin{center}
        \includegraphics[width=3.1in]{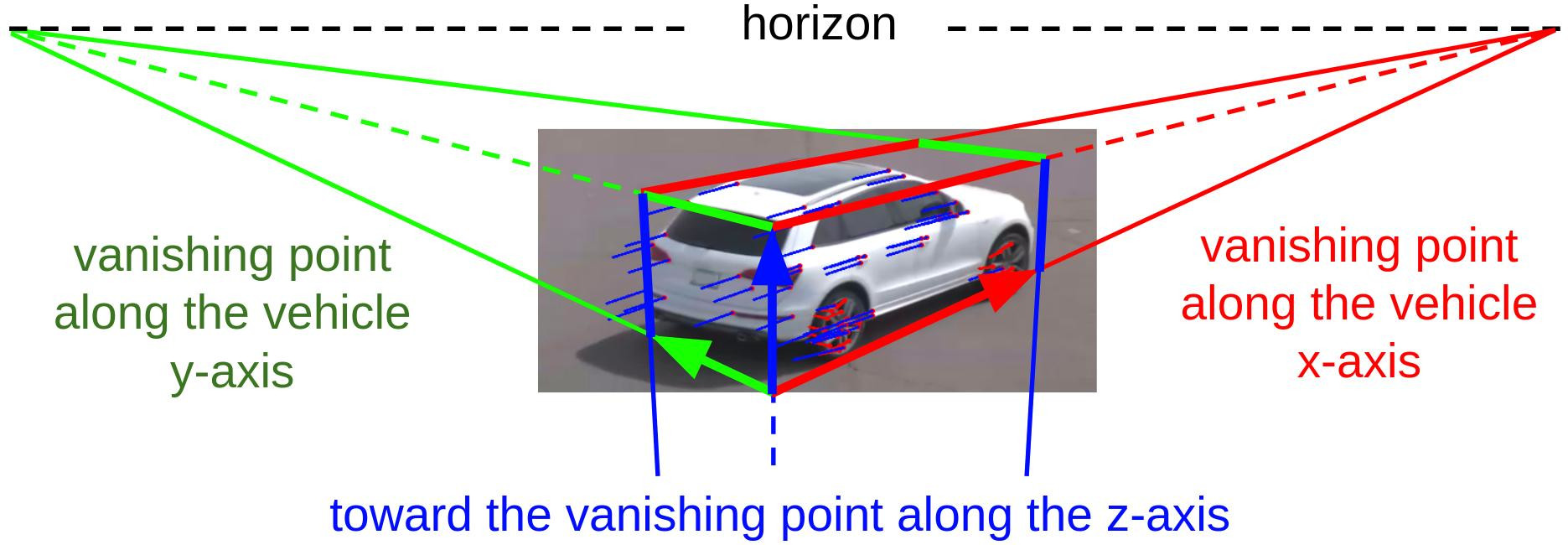}
    \caption{An illustration of 3D bounding box. The optical flow vector inliers are shown as the thin blue lines (on the vehicle body) and outliers are shown as the thin red lines (on the wheels).}
    \label{fig:bb3d}
    \end{center}
\vspace{-0.28in}
\end{figure}

\textbf{(4) Vehicle Localization}: The system runs RANSAC \cite{fischler1981random} on the computed optical flow vectors obtained in the previous vehicle tracking stage to select those vectors that meet at the same vanishing point on the horizon line (which is computed from the camera calibration results), as shown in Fig. \ref{fig:bb3d}. Because a vehicle rarely moves backward on the road, the vehicle heading is determined by the line from the center of its 2D bounding box to this vanishing point. The vehicle has its XYZ coordinate reference frame where the x-axis points to the vehicle's forward, and the y-axis points to its left. We assume that the vehicle is always on the ground, \textit{i.e.}, its z-axis is pointing up relative to the ground surface. With the center of its 2D bounding box as its temporary location on the image, the transformation $T$, and its heading, the system computes the other two vanishing points corresponding to the y-axis and z-axis of the vehicle. Finally, using all three vanishing points, the 3D bounding box of a vehicle is computed from the contour of its segmentation mask using the tangent line method \cite{dubska2014automatic} (illustrated in Fig. \ref{fig:bb3d}). We made several improvements in implementation details to handle a few particular viewing angles not considered in \cite{dubska2014automatic}, and we also made adjustments on the computed 3D bounding box using empirical results to accommodate vehicles without ``boxy'' shapes. Besides, we use the recognized vehicle type and prior knowledge of the vehicle dimensions for different vehicle types to adjust the 3D bounding box dimensions. Additionally, the 3D bounding box is not calculated if certain occlusion conditions are detected using the 2D bounding box overlap and the size of the mask. Finally, the center of this 3D bounding box's bottom surface is the vehicle's location on the image. Again, with the transformation $T$, the location of the vehicle in the world frame is obtained.

The heading calculation may fail in a few cases: (a) when the vehicle stops, (b) when the vehicle is far away with only small motion on the image, or (c) when the RANSAC fails. In these cases, the heading is inferred from the accumulated motion on several previous video images by assuming the vehicle travels in a straight line within a short amount of time. Moreover, we use a neural network similar to \cite{juranek2015real} to predict the heading angle of a vehicle on the image as a backup. It is trained on a dataset of images patches with correct heading calculated by the optical flow based method. However, this neural network method is generally slower, less accurate and less robust than the optical flow method.

\textbf{(5) Vehicle Velocity Measurement}: The system first averages the length of the inlier vectors of the RANSAC results obtained in the previous step and use this length as the distance of vehicle movement from the current image and the previous image. Next, it also obtains the corresponding distance in the world frame using the vehicle's location, heading, and the transformation $T$. After that, the system uses the linked list of associated instances to aggregate these distances calculated from previous image pairs in sequence until the total distance exceeds a threshold or up to a certain amount of steps (5 m or 30 steps in our implementation). Finally, the velocity is calculated from this aggregated distance, the number of frame pairs, and the frame interval time. The direction of the velocity is the same as the heading. 

\textbf{(6) Vehicle State Estimation}: Given the location and the velocity of a vehicle, the system runs a Kalman filter with states $\mathbf{x} = (x, y, z, \dot{x}, \dot{y}, \dot{z})$ and linear 6DOF rigid body kinematics to estimate the vehicle states in the 3D world. The process and observation noise covariance matrices are empirically determined. For a 2D map, $z$ and $\dot{z}$ are always zero. The prediction step of the state estimation will keep running for a few iterations when the detection of this vehicle fails. Once it is detected again, a detected instance can be re-associated to this vehicle. The heading and 3D bounding box dimension is also smoothed using a running average on previous video images. Finally, a record of the location, velocity, heading, 3D bounding box points, and the vehicle type is created as the output of this pipeline.

\subsection{Vehicle Shape Reconstruction}

\begin{table*}[]
\vspace{0.07in}
\centering
\begin{tabular}{|c|c|c|c|c|c|c|c|c|c|c|c|c|c|c|}
\hline
Videos   & MOTA   & MODA   & MME   & FP  & FN    & \#Objects & \#Images & \#Vehicles & \#IDE & \#TO & \#PO & MT  & ML & Resolution \\ \hline
Track 1A & 96.2\% & 98.1\% & 220   & 20  & 3,802 & 95,227    & 17,891   & 286        & 22    & 7    & 112  & 271 & 5  & 720p       \\ \hline
Track 1B & 90.9\% & 92.4\% & 1,218 & 330 & 5,670 & 79,210    & 17,912   & 225        & 40    & 10   & 109  & 207 & 6  & 720p       \\ \hline
Track 2  & 95.5\% & 96.3\% & 65    & 0   & 496   & 12,346    & 25,458   & 80         & 1     & 0    & 2    & 75  & 2  & 1080p      \\ \hline
\end{tabular}
\caption{Tracking results.}
\label{tb:tracking}
\vspace{-0.2in}
\end{table*}

\begin{figure}[]
\vspace{0.07in}
    % \centering
    \begin{center}
        \includegraphics[width=3.3in]{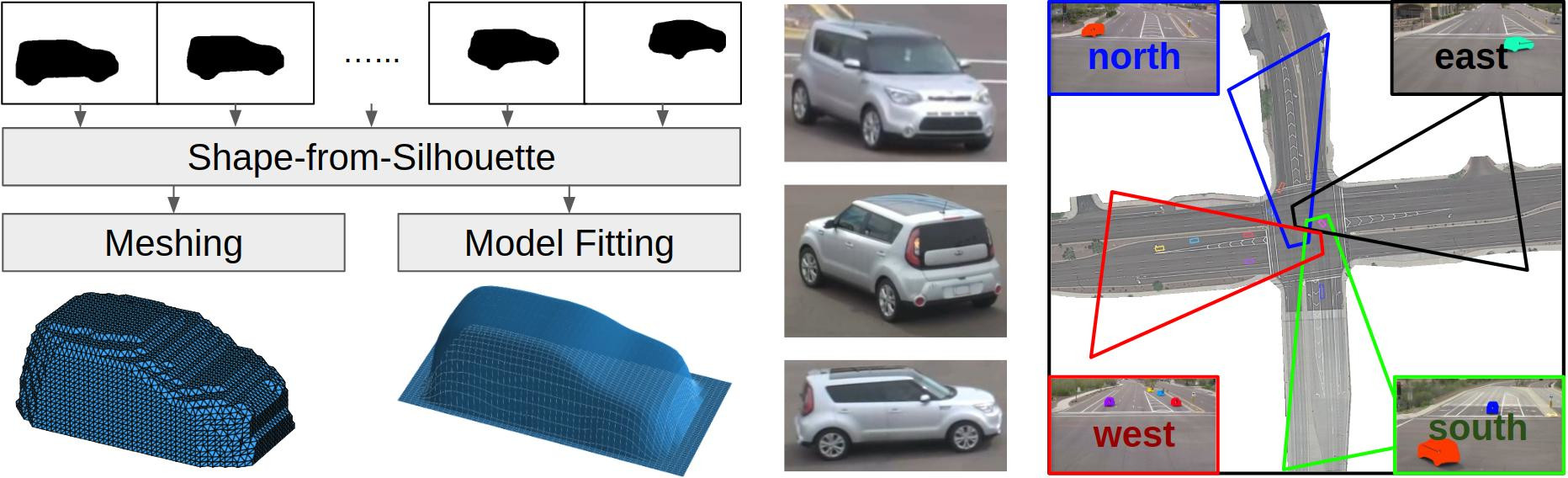}
    \caption{The vehicle shape reconstruction pipeline with the reconstructed 3D shape of an example vehicle (left), three images of the vehicle (middle), and the intersection with four cameras providing the videos (right).}
    \label{fig:reconstruction}
    \end{center}
\vspace{-0.3in}
\end{figure}

We applied the tracking pipeline on the traffic monitoring videos obtained from four cameras pointing to the four directions of an intersection, as shown in Fig. \ref{fig:reconstruction} (right). These videos allow us to observe the same vehicle traveling through the intersection from multiple viewing angles. Given a sequence of vehicle locations, 3D bounding boxes, and segmentation masks from multiple images, we compute the vehicle's visual hull using the shape-from-silhouette method \cite{laurentini1994visual}. Specifically, we first initialize a rectangular cuboid of voxels using the 3D bounding box. Then we carve those voxels that cannot be projected onto any mask on any views. After that, we further process the remaining voxels using the symmetry property along the vehicle's x-axis. The voxels can be converted to a mesh using the marching cubes algorithm \cite{lorensen1987marching} for visualization, as in Fig. \ref{fig:reconstruction} (left). The voxels can also be converted to a 2D histogram by ignoring the details on the bottom side. Specifically, for each histogram bin at $(x, y)$ in the vehicle's own XYZ coordinate frame, the histogram value is the maximum of the $z$ coordinates of all remaining voxels with this $(x, y)$ coordinates. Similarly, a 2D histogram can be converted back to voxels. We further resample the 2D histogram to a fixed size of n-by-m bins using bilinear interpolation. The result histogram is denoted as a n-by-m matrix $H$, or flattened as a n*m dimensional vector $\mathbf{h}$. In our implementation, $n=m=50$.

Our objective is to reconstruct the vehicle shape and representing it in a fixed-sized data structure. Here $H$ can be a candidate. However, it usually differs from the actual vehicle shape due to the limited view angles in the voxel carving process and errors in the localization results. To solve this problem, we construct a shape prior model from 80 different 3D CAD vehicle models and fit the model to the reconstructed histogram $H$ by the following procedures. 

\textbf{Step (1)}: For each 3D CAD model, we converted it to a histogram using an algorithm similar to the one that converts voxels to a histogram. The generated histogram was resampled to n-by-m, and flattened to a n*m vector (denoted as the model vector $\{\mathbf{u}_i\}$, $1 \le i \le 80$).

\textbf{Step (2)}: With all 80 model vectors, we run Principal Component Analysis (PCA) to reduce their dimension from n*m to 20. After this step we obtained 20 principle component vectors (denoted as a n*m-by-20 matrix $S$). The vector set $\{\mathbf{u}_i\}$ and matrix $S$ are called the \textbf{vehicle shape prior}, similar to the shape prior models in shape analysis and multiple view reconstruction \cite{cootes1995active}\cite{engelmann2017samp}\cite{chhaya2016monocular}.

\textbf{Step (3)}: We projected the reconstructed histogram $\mathbf{h}$ to the column space spanned by $S$ by solving the following least-square problem:
$$\underset{\mathbf{v}}{\arg\min} || \mathbf{h} - S\mathbf{v} || + \lambda || \mathbf{v} - \mathbf{t}||.$$
Here the last term is a regularizer, $\mathbf{t}$ is a template vector for the type of the reconstructed vehicle, and it is computed by averaging the subset of $\{\mathbf{u}_i\}$ with the same type. For example, if $\mathbf{h}$ is reconstructed from a vehicle that is recognized as a ``sedan'', $\mathbf{t}$ is the template vector of ``sedan'', which is computed by averaging of those $\mathbf{u}_i$ derived from 3D models of sedans. Moreover, $\mathbf{t}$ is also used to represent those vehicles whose shapes cannot be reconstructed due to occlusion.

Finally, the vector $\mathbf{v}$ is the output of this pipeline as the shape representation of the reconstructed vehicle. Given $\mathbf{v}$ and $S$, an approximated histogram representation of the vehicle shape can be recovered by $\mathbf{\hat{h}} = S\mathbf{v}$. This histogram can be further converted to voxels or a mesh. The texture of the vehicle 3D model is not reconstructed for anonymity.

\subsection{Map Reconstruction}

We constructed the 2D map using a satellite image at the place where the camera is mounted. Many online map services (such as Google Maps) offer satellite images. The rows and columns of the image are usually already aligned to the east and the south. We also calculate the scale factor between the map pixel and the metric unit using two points with known actual distance in the metric unit (which can be measured using the online map service tools or in the world).

For the 3D map, we flew a survey-grade drone (DJI Phantom Pro with RTK) on the site, ran a 3D reconstruction software (Pix4D mapper) to obtain a point cloud from the drone images, and then processed the point cloud to a 3D mesh map. We also calibrated this mesh map to align its axes to east-north-up and recover the actual scale.

Additionally, we chose a reference point for both types of maps and obtained its longitude, latitude, and height above the geodesic ellipsoid using the online map service or a hand-held GPS receiver device on site. Then, we set up the transformation between the map reference frame to the WGS84 reference frame so that we can compare our localization results with GPS measurements.

\subsection{Traffic Scene Visualization and Replay}

To replay a traffic scene captured by the cameras, we built two visualizers, one using the 2D map and the other using the 3D map, as shown in Fig. \ref{fig:overview}. Here, a traffic scene is defined as the collection of the road environment (\textit{i.e.}, the map) and vehicles captured by a specific camera within a certain period (\textit{i.e.}, the tracking results). Both visualizers transform the vehicle states to the map frame using the calibration results and animate the vehicle movement. We use a template 3D mesh model for each vehicle type or the reconstructed vehicle models for the 3D animation. The size of the 3D model is scaled to fit the vehicle 3D bounding box. Besides, during the replay, the user can modify the speed of one specific vehicle, and the visualizer can ``re-simulate'' this vehicle from the modified states following the recorded trajectory while keeping replaying other vehicles.

\section{Empirical Evaluation}

\begin{figure}[]
    % \centering
    \begin{center}
        \includegraphics[width=3.35in]{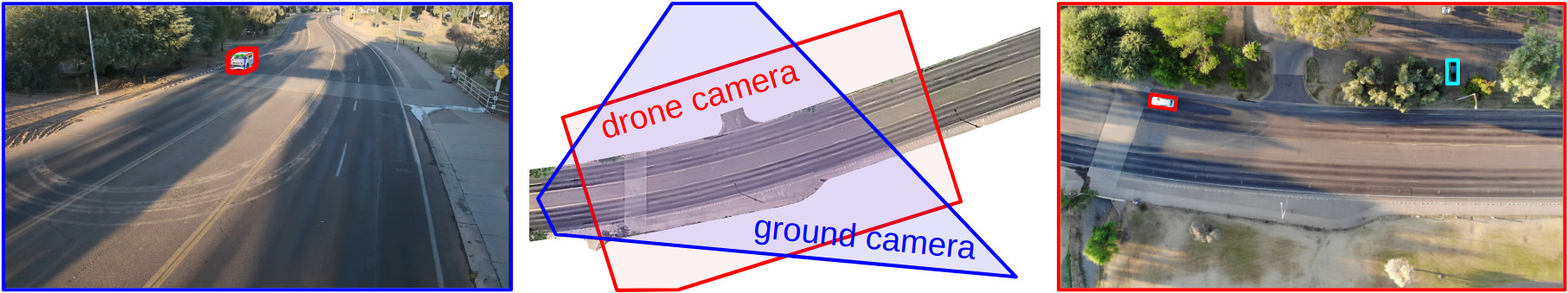} with
    \caption{Example images taken by the ground camera on the road infrastructure (left) and the drone at a height of 80 meters (right) at the second site. The coverage of both cameras are shown on the map (middle).}
    \label{fig:drone}
    \end{center}
\vspace{-0.15in}
\end{figure}

We obtained traffic monitoring videos from two sites. The first one is an intersection with four cameras pointing to its four directions (the same intersection in \cite{wishart2020driving}), which is the one shown in Fig. \ref{fig:reconstruction} in section III.C. The second site is a local road segment with one camera, shown in Fig. \ref{fig:drone}.

First, we evaluate the vehicle type recognition performance. The recognizer is trained with 10,200 images and tested with 883 images. All images are cropped from the videos recorded by the four cameras at the first site. The overall accuracy is 84\%. The majority of the wrong predictions are among the following type pairs: (SUV, sedan), (SUV, minivan), (sedan, coupe). Prediction errors are more frequent when only the frontal side or the rear side of the vehicle is visible, \textit{i.e.}, when the vehicle is driving directly towards the camera or away from the camera.

\begin{figure*}[]
\vspace{0.07in}
    % \centering
    \begin{center}
        \includegraphics[width=6.8in]{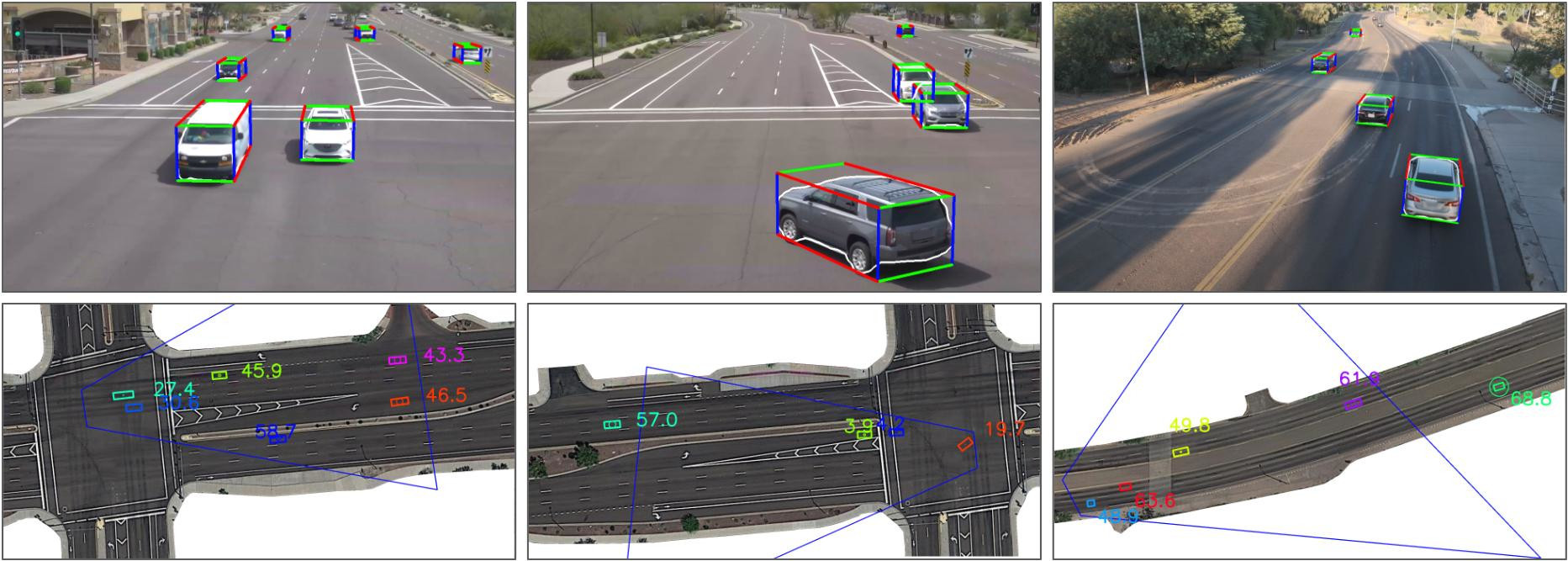}
    \caption{Examples of reconstructed traffic scenes. The first row shows the original video with vehicle 3D bounding boxes. The second row shows on the map with the vehicle location with uncertainty range (the rectangles and the circles on them) and speed in km/h (the numbers adjacent to the rectangles).}
    \label{fig:scene}
    \end{center}
\vspace{-0.1in}
\end{figure*}

\begin{figure*}[]
    % \centering
    \begin{center}
        \includegraphics[width=6.9in]{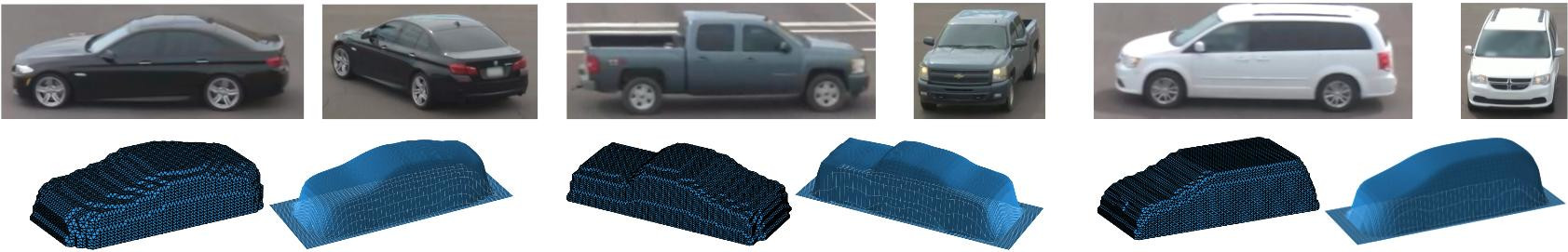}
    \caption{Examples of reconstructed vehicles with voxel representations and histogram shapes.}
    \label{fig:shapes}
    \end{center}
\vspace{-0.3in}
\end{figure*}

Second, we evaluate the tracking performance on the 2D images with two video tracks from the first site (the eastbound and the southbound) and one video track from the second site. The results are shown in TABLE \ref{tb:tracking}, mostly following the metrics in \cite{stiefelhagen2006clear}. Here ``\#Objects'', ``\#Images'' and ``\#Vehicles'' mean the total number of objects, images, and vehicles in the video track. ``\#IDE'' means the number of vehicles that have tracking ID errors. ``\#TO'' means ``total occlusions'', which is the number of vehicles that are occluded by other vehicles in at least one image such that more than 80\% of the vehicle is not visible. Generally, the detector fails to detect them and the tracker needs to re-associate it later. If a vehicle is partially occluded but still detected, it is counted in ``\#PO'', which means ``partial occlusions''. Typically, the traffic monitoring cameras are mounted at strategically chosen places to minimize occlusion. ``\#MT'' is the number vehicles that are tracked more than 80\% of the life span (\textit{i.e.}, ``mostly tracked''). ``\#ML'' is the number vehicles that are tracked less than 20\% of the life span (\textit{i.e.}, ``mostly lost''). Our system is able to track most detected vehicles that are not completely occluded.

\begin{table}[]
\centering
\begin{tabular}{|c|c|c|c|c|c|}
\hline
Video    & \begin{tabular}[c]{@{}c@{}}L-Diff \\ (m)\end{tabular} & \begin{tabular}[c]{@{}c@{}}V-Diff \\ (m/s)\end{tabular} & \begin{tabular}[c]{@{}c@{}}\#Vehicles \\ (w/ Ref) \end{tabular} & \begin{tabular}[c]{@{}c@{}}Coverage\\ (m)\end{tabular} & \begin{tabular}[c]{@{}c@{}}Ref \\ Device\end{tabular} \\ \hline
Track 1A & 2.05                                                  & 1.01                                                    & 1          & 25 $\sim$120                                        & GPS                                                   \\ \hline
Track 1B & 1.57                                                  & 0.69                                                    & 1          & 25 $\sim$120                                        & GPS                                                   \\ \hline
Track 2  & 1.68                                                  & 1.47                                                    & 69         & 15 $\sim$110                                        & Drone                                                 \\ \hline
\end{tabular}
\caption{Localization and speed measurement results.}
\label{tb:location}
\vspace{-0.25in}
\end{table}

Third, we quantitatively evaluate the vehicle localization and velocity measurement performance in the 3D world frame using two different types of references. For the first site, we drove a vehicle with a differential GPS receiver through the intersection. For the second site, we flew a drone and capture videos from 80 meters above the road. We processed the drone videos to obtain the vehicle location and velocity using a method similar to \cite{zhan2019interaction}. We also compared the location measurements between the GPS and the drone using another drone video track that captures the movement of a vehicle with a differential GPS receiver. The location and velocity measurements accuracy of both types of reference devices are within 1 m and 1 m/s respectively. Our results are shown in TABLE \ref{tb:location}, where ``L-Diff'' and ``V-Diff'' mean differences in location and velocity between our results and the reference measurements. Only the vehicles that are correctly tracked and those with corresponding reference measurements are evaluated, as shown in the ``\#Vehicles'' columns. The ``Coverage'' column shows the minimum distance and the maximum distance between the measured vehicles and the camera. Note that ``L-Diff'' actually varies with the distance between the vehicle and the camera, and typically it is smaller when the vehicle is close to the camera. For example, the average value of ``L-Diff'' is 0.79 m within 50 m to the camera in track 2, which is less than the average ``L-Diff'' value in the whole range. Also, ``L-Diff'' is not the same in the longitudinal direction (\textit{i.e.}, the camera pointing direction) and the lateral direction (\textit{i.e.}, perpendicular to the camera pointing direction). For example, in track 2, the average lateral location difference is just 0.22 m but the average longitudinal location difference is 1.52 m. At certain view angles, e.g., when the vehicle is driving directly towards the camera, the 3D bounding box is inaccurate. Moreover, the road surface is not perfectly flat, which can cause errors in the conversion of image coordinates to the world coordinates when the 2D map is used. Converting world coordinates to GPS coordinates may also introduce small errors.

Fourth, we show the qualitative results in Fig. \ref{fig:scene}, Fig. \ref{fig:shapes}, and the supplemental videos\footnote{https://github.com/duolu/CAROM}. We also obtained positive feedback from Maricopa County DOT and the Institute of Automated Mobility \cite{IAM} in Arizona.

At last, we discuss the limitations of the current system and possible future improvements. Currently, both the 3D bounding box calculation and the shape-from-silhouette algorithm require a complete segmentation mask. Inaccurate results are generated for vehicles with partial occlusion. We plan to collect a large-scale dataset and train a neural network pose estimator that can work robustly under partial occlusion. We also aim to train a neural network to directly predict the vehicle shape vectors and its pose jointly from a single image using the current vehicle reconstruction results as training data, so that the 3D shape of a vehicle can be directly obtained at every frame. This will enable us to develop a model-based tracking algorithm to increase the processing speed and improve the robustness. Besides, pedestrians, cyclists, and other types of traffic participants are not reconstructed in our current implementation, and we want to calculate ``bounding cylinders'' for these moving objects with non-boxy shapes. Finally, we are working on calculating safety metrics \cite{wishart2020driving} from our tracking results as an application.

\section{Conclusions}

In this paper, we present CAROM, a vehicle localization and traffic scene reconstruction framework using videos taken by monocular cameras mounted on road infrastructures, which achieves promising results for vehicle localization and velocity measurement. Still, CAROM is in its early stage with limitations in robustness and efficiency. With further development, we hope it can be deployed together with traffic monitoring cameras on the roadside infrastructure in the future, to allow jurisdictional authorities and AVs on the road gain better awareness of the traffic situation.

%The current implementation records and processes videos in batches. Our future work includes deploying it on public roads together with the cameras to assist regulators in local DOTs and regulate self-driving vehicles. 
%safer on the road.

\bibliographystyle{IEEEtran}
\bibliography{references}

\end{document}